\documentclass[11pt,a4paper]{article}

\usepackage{authblk}

\usepackage{amsmath,amsfonts}
\usepackage{algorithmic}
\usepackage{algorithm}
\usepackage{array}
\usepackage[caption=false,font=normalsize,labelfont=sf,textfont=sf]{subfig}
\usepackage{textcomp}
\usepackage{stfloats}
\usepackage{url}
\usepackage{verbatim}
\usepackage{graphicx}
\usepackage{cite}

\usepackage[pagebackref,breaklinks,colorlinks]{hyperref}

\usepackage{booktabs} 
\usepackage{svg}
\usepackage{multirow}
\usepackage{multicol}
\usepackage{orcidlink}
\newcommand{\first}[1]{\textbf{#1}}
\newcommand{\second}[1]{\underline{#1}}
\newcommand{\third}[1]{\underline{\underline{#1}}}
\def\NuestroMetodo{PO-GUISE+}

\begin{document}

\title{PO-GUISE+: Pose and object guided transformer token selection for efficient driver action recognition}

\author[1]{Ricardo Pizarro}
\author[2]{Roberto Valle}
\author[1]{Rafael Barea}
\author[3]{Jos\'e M. Buenaposada}
\author[2]{Luis Baumela}
\author[1]{Luis Miguel Bergasa}
\affil[1]{Universidad de Alcal\'a de Henares, \texttt{\{ricardo.pizarroc, rafael.barea, luism.bergasa\}@uah.es}}
\affil[2]{Universidad Polit\'ecnica de Madrid, \texttt{\{rvalle,lbaumela\}@fi.upm.es}}
\affil[3]{Universidad Rey Juan Carlos, \texttt{josemiguel.buenaposada@urjc.es}}

\maketitle

\begin{abstract}
We address the task of identifying distracted driving by analyzing in-car videos using efficient transformers. Although transformer models have achieved outstanding performance in human action recognition tasks, their high computational costs limit their application onboard a vehicle. We introduce \NuestroMetodo{}, a multi-task video transformer that, given an input clip, predicts the distracted driving action, the driver's pose, and the interacting object. Our enhanced features for token selection are specifically adapted to driver actions by leveraging information about object interaction and the driver’s pose. With \NuestroMetodo{}, we significantly reduce the model's computational demands while maintaining or improving baseline accuracy across various computational budgets. Additionally, to evaluate our model's performance in real-world scenarios, we have developed benchmarks on a Jetson computing platform, demonstrating its effectiveness across different configurations and computational budgets. Our model outperforms current state-of-the-art results on the Drive\&Act, 100-Driver, and 3MDAD datasets, while having superior efficiency compared to existing video transformer-based methods.
\\
\vspace{1em}
\end{abstract}

\section{Introduction}
Driver distraction is a significant concern for road safety. According to Eurostat statistics, 20,400 people were killed in road accidents in the EU in 2023~\cite{Transport_2024}.
The exact number of road accidents resulting from distracted drivers remains unclear. Austrian crash data indicate that distraction and inattention were responsible for 29\% of injury crashes and 25\% of fatal crashes in 2022\cite{STATISTIK_2024,distr_2023}.
While estimates indicate that 5-25\% of European accidents result from driver distraction —encompassing activities such as cellphone or GPS usage, eating, smoking, fatigue or stress— research conducted in real-life driving scenarios unveils that 68.3\% of crashes exhibit some signs of discernible distraction~\cite{distr_eu_2022}. The consequences of these individual incidents cascade throughout the broader network. An initial crash frequently triggers dangerous traffic conditions, such as sudden congestion. Additionally, these traffic characteristics are directly linked to an increased likelihood of subsequent crash occurrences~\cite{roshandel2015impact}. Therefore, robust in-vehicle Driver Monitoring Systems (DMS) are not merely a feature for individual vehicle safety, but a foundational technology to improve the predictability, reliability, and safety of the entire transportation ecosystem.

To fulfill this role, the core function of a driver monitoring system is to identify whether a driver is focused or distracted. These systems must determine whether the driver is attentively observing the road or engaging in other activities, such as glancing around the cabin or using a cellphone. This requires deploying advanced algorithms and machine learning techniques that can accurately analyze data from sensors and cameras.

In this paper, we study the problem of recognizing different situations of distracted driving based on in-car video analysis.

Driver distraction detection is intimately related to the problem of human action recognition. In human action recognition we have seen a transition from methods using CNNs~\cite{simonyan2014twostream, wang2016tsn, carreira2017i3d} and 3D-CNNs~\cite{feichtenhofer2016spatiotemporal, tran2018r21d, lin2019tsm} or a mixture of both~\cite{carreira2017i3d}
to transformers~\cite{chen2023semantic, chen2023efficient, reilly2024just}. Recent large-scale datasets combined with self-supervised techniques have enabled the creation of video transformer models that achieve top accuracy in the human action recognition domain~\cite{wang2023videomaev2}. Many of these advances have been applied to the detection of driver distraction~\cite{hongdetecting2022,tan2024survey}, leading to significant improvements in this area.  Currently, video transformers achieve the best results in the driver distraction detection task~\cite{peng2022transdarc,
pizarro2024drvmon}, but their high computational cost poses a significant challenge for practical implementation in onboard vehicle systems. This cost is primarily due to their quadratic computational complexity, which significantly escalates computational demands as the number of spatio-temporal tokens increases.

\begin{figure}
    \centering  
    \includegraphics[width=\columnwidth]{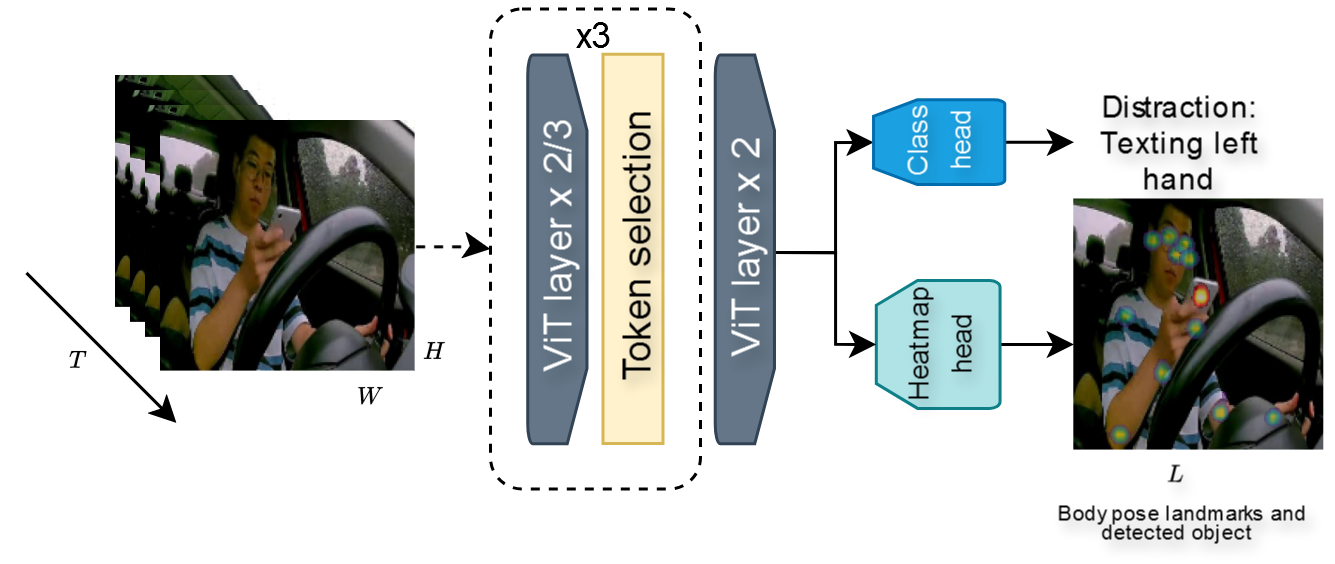}

    \caption{Driver distraction recognition task overview. An input video clip is processed by a ViT transformer improved by \NuestroMetodo{} to increase efficiency. The resulting output is a distraction classification, and heatmap locations corresponding to the driver and interacting object. }  
    \label{fig:overview_task}  
\end{figure} 

One method to achieve a lower computational cost with a trade-off between accuracy and efficiency is token selection. In this approach, a percentage of tokens are discarded at certain blocks within the transformer model, thereby reducing the total number of tokens the model needs to process. Popular techniques include Top-K~\cite{haurum2023tokens}, where token selection is guided by retaining the K tokens with the highest attention to the class token or merging similar tokens~\cite{bolya2023tome}. Our recent technique PO-GUISE~\cite{pizarro2024poguise} leverages human pose and action information to improve token selection in general human action recognition contexts. A limitation of this technique is that it overlooks the interactions with objects, which is crucial in driver distraction recognition. This results in suboptimal performance, especially when lowering the computational budget, where the drop in accuracy is more noticeable.

In this paper we extend previous works~\cite{pizarro2024drvmon,pizarro2024poguise} and introduce an enhanced token selection method for transformer models that uniquely leverages the intricacies of driver action recognition. Our approach integrates semantic information from actions, object interaction locations, and human poses to refine token selection and reduce computational demands while maintaining accuracy. Our technique is implemented in two ViT-based baselines: VideoMAEv2~\cite{wang2023videomaev2} and InternVideo2~\cite{wang2024internvideo2}, which are transformer architectures pre-trained with a self-supervised strategy and refined with a large human action database. Our method, referred to as \NuestroMetodo{}, is trained in a multi-task fashion using only video clips as input. These clips are converted into spatio-temporal visual tokens and are processed alongside heatmap tokens. Our representation of heatmap tokens offers the flexibility to extend the traditional representation of heatmaps into features that are able to represent the temporal changes in human poses and the locations of relevant objects within the scene. Specifically, we generate a motion heatmap for each landmark and another for the interacting object, representing the movement across all frames in each clip (see Fig.~\ref{fig:hm_gt}). Fig.~\ref{fig:overview_task} provides an overview of the driver action recognition task.

The token selection method consists of a two-step process. The first step is an enhanced token pruning which discards spatio-temporal visual tokens referred to as \emph{video tokens}, that do not sufficiently attend to relevant information about the distraction class, driver pose, and interacting objects. 
In the second step, the merging method summarizes the pruned tokens by averaging similar dropped tokens. This token selection module is integrated at various stages of a transformer model, with the number of selected tokens controlled by a continuous \textit{token keep rate}. The features used to guide the pruning step are crucial to the accuracy of the model, especially at low \textit{token keep rates}, as the model has fewer resources to process the entire video. Fig.~\ref{fig:token_sel} shows an example of the use of only distraction class and pose to guide token selection compared to our improved method, \NuestroMetodo{}.

As far as we know, we are the first to integrate distraction recognition, human pose and object location estimation in a multi-task video transformer, that does not require an external pose or object detector. This innovation aims to reduce the computational cost of the model while uniquely adapting a token selection method for driver action recognition using only video input. To thoroughly evaluate our model's performance in a realistic scenario, we have developed an extensive set of benchmarks involving a Jetson computing platform to estimate both the accuracy and latency across different computational costs and model sizes.

In summary, the contributions of our work are as follows:
\begin{itemize}
    \item We propose a token selection method adapted to the task of driving action recognition by leveraging the interacting object location, driver's pose, and distraction class information. It improves the trade-off between efficiency and accuracy compared to other methods from the state-of-the-art, particularly at the most efficient low \textit{token keep rates}.
    \item Our multi-task model provides distraction classification, driver's pose, and object location simultaneously from a video clip without relying on external detectors.
    \item Our method surpasses the state-of-the-art in multiple driver action recognition datasets while being much more efficient than current video transformer-based methods.
    \item We provide a comprehensive evaluation at different points in the accuracy-performance trade-off curve to demonstrate practical real-world usage on a Jetson platform.
\end{itemize}

\section{Related work}
\label{sec:related}
This section discusses the specifics of driver action recognition and provides an overview of techniques to overcome the computational requirements of transformers.
Recognizing actions in videos requires considering variations in the actors' positions and poses, their movements, as well as their interactions with objects.

{\setlength{\parindent}{0cm}
\textbf{Driver action recognition.}} 
Deep learning models have been widely employed for the task of human action recognition in videos. Various methods have been developed to extend these models to the temporal domain. An approach involves incorporating motion information by performing convolutions in both spatial and temporal dimensions using 3D Convolutional Neural Networks (3D CNNs)~\cite{feichtenhofer2016spatiotemporal}. A more effective method for introducing temporal context is the application of transformers on top of features extracted by 3D CNNs~\cite{kalfaoglu2020late}. Other approaches focus on learning motion-driven visual tempo to better capture temporal dynamics~\cite{liumotion2022}. Alternatively, Vision-Language Models (VLMs) have been explored to identify distracted behaviors by using natural language descriptions~\cite{hasan2024visionlanguage}.
The use of probability maps or heatmaps to locate body keypoints has also been proven to be highly discriminative in human action recognition~\cite{choutas2018potion, liu2018recognizing, yan2019pa3d, shah2022jmrn, ahn2023star,sangwon2023cross,zhang2024pgvt}.

Although human action recognition is a well-studied problem, effectively detecting driver distraction presents unique challenges. It requires a diverse and extensive dataset to accurately capture the wide range of behaviors that might occur in real driving situations. Various sensors and modalities have been used to address this challenge~\cite{tan2024survey}. Vision-based sensors are crucial for analyzing driver movements to detect distractions and facilitate real-world applications. Notable datasets such as Drive\&Act~\cite{martin2019driveact}, DMD~\cite{ortega2020dmd}, 3MDAD~\cite{jegham20203mdad}, and ``100 drivers"~\cite{wang2023100driver} cover a broad spectrum of driver interactions, actions, and camera perspectives.
%
The Drive\&Act dataset is particularly prevalent within this research area due to its large extent.
The standard way of evaluation is to use the front-top camera with the Near Infrared (NIR) modality. Early methods~\cite{martin2019driveact} adopted this methodology as a baseline, including fine-tuned two-stream 3D CNNs 
pre-trained in a popular action recognition dataset,  Kinetics~\cite{carreira2017i3d}, (e.g., I3D~\cite{carreira2017i3d}) and also skeleton-based methods~\cite{martin2019driveact}. A more recent work, CTA-Net\cite{wharton2021ctanet}, uses CNNs, self-attention layers, and LSTMs. 
Recent works have focused on the use of advanced vision transformers. The first of these is TransDARC~\cite{peng2022transdarc}, which leverages the Video Swin Transformer~\cite{liu2022videoswing} paired with feature augmentation strategies to distill video features to generate new training samples based on the rareness of the class. CoViT~\cite{licovit2024} proposed a hybrid model that fuses a CNN with a Vision Transformer for efficient single-frame distracted driver classification. DRVMonVM~\cite{pizarro2024drvmon} also leverages state-of-the-art vision data augmentation techniques coupled with VideoMAEv2, a vision transformer model pre-trained in a self-supervised manner on a giant-scale dataset.

Similar lines of work can be observed on the 3MDAD dataset~\cite{jegham20203mdad}. 
Recent methods use lightweight transformers, such as
Feature Pyramid Vision Transformer~\cite{wang2023FPT}
or LW-transformer~\cite{mohammed2024}, within a teacher-student framework. %
Other contemporary works, DACNet~\cite{su2023DADCNet} and MIFI~\cite{kuang2023MIFI}, explore the multi-view (DACNet, MIFI) and the multi-modal approaches (DACNet).
An alternative modality to RGB or Infrared (IR) images is the use of 2D or 3D body-pose in the form of keypoints location, some times given as heatmaps and others as a graph~\cite{li2023oasar,holzbock2022stmlp,li2024smoms}. 
The main drawback of these models
is that they require an external model to detect the keypoints in the image.

{\setlength{\parindent}{0cm}
\textbf{Video Transformers computational requirements}.} 
A limitation of previous works~\cite{peng2022transdarc,
pizarro2024drvmon} is the quadratic complexity in the number of tokens of transformers. This is a fundamental constraint for the real-world deployment of these models. It can be addressed in different ways. 
One is to factorize the attention along the spatial and temporal dimensions~\cite{arnab2021vivit}.
Another way is to merge similar tokens into new ones~\cite{bolya2023tome} or pruning the less promising tokens as measured by the attention to different objectives ~\cite{liang2022evit, ma2022ppt,chen2023efficient}.
Our recent technique, PO-GUISE~\cite{pizarro2024poguise}, improves upon these methods by adding the body pose and guiding pruning by the attention to tokens that relate to both the class and body pose tokens. Although this method is effective for general human action recognition, it overlooks the role of object interactions, missing critical information, especially at low computational budgets.

\textbf{Our proposal.}
We use a multi-task strategy to adapt general human action recognition token selection to the specific task of driver action recognition by estimating human pose, object location heatmaps, and driver actions. This approach enhances previous methods, by maintaining accuracy while operating under a low computational budget. Additionally, our proposed pose-object guided video token selection method improves the accuracy of the baseline model and reduces the computation by 30\% in the default settings. It sets new state-of-the-art results across multiple driver action recognition datasets (see Tables~\ref{tab:driveactsota},~\ref{tab:hundredsota},~\ref{tab:3mdadsota}). Furthermore, we extensively benchmark different settings on a Jetson computing platform to evaluate our model in a real-world situation.

\section{Pose\&Object-guided token selection for multi-task video transformers (\NuestroMetodo)}
\label{sec:method}
\begin{figure*}
\centerline{\includegraphics[width=\linewidth]{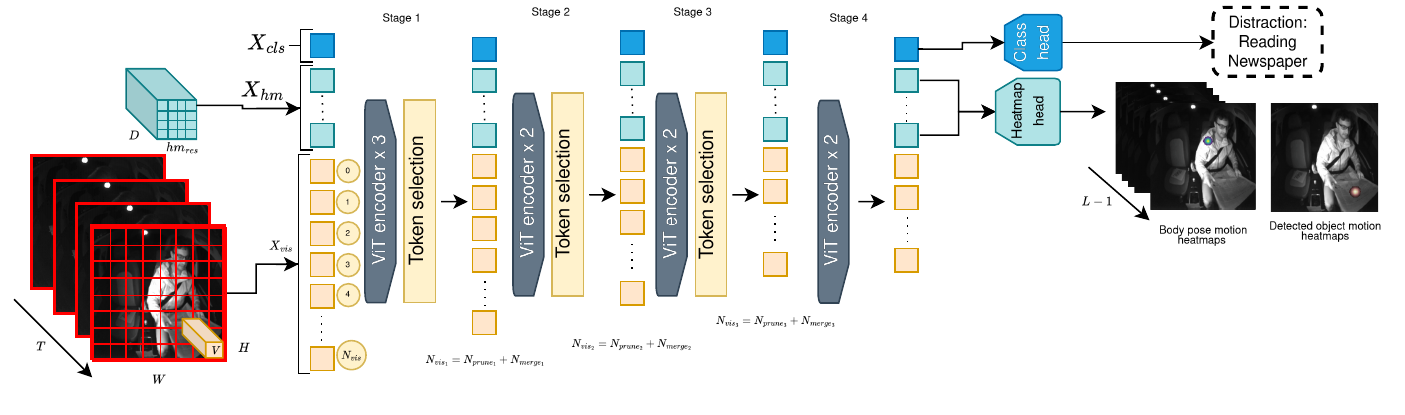}}
\caption{An input clip is tokenized into visual tokens ($X_{vis}$) and processed by a Vision Transformer (ViT) encoder, which also includes learnable class tokens ($X_{cls}$) and heatmap tokens ($X_{hm}$). In the first three stages, the enhanced token selection module reduces the number of tokens in $X_{vis}$, lowering the computational cost of the model. The output from a video clip includes the driver distraction class and the corresponding heatmaps representing the body pose and interacting object.}
\label{fig:modeldiag}
\end{figure*}

Our approach employs a pre-trained ViT-based video transformer encoder, using the top-performing DRVMon-VM~\cite{pizarro2024drvmon} as a starting point. We address computational complexity by extending the PO-GUISE~\cite{pizarro2024poguise} framework through a unified multi-task architecture tailored for the driver monitoring domain. This architecture enhances the ViT backbone with pose and action recognition tasks to facilitate efficient token selection. Notably, \NuestroMetodo{} uniquely incorporates interacting-object heatmaps alongside skeletal pose to refine token pruning, ensuring the model focuses on critical driver-object interactions. While we rely on ViTPose~\cite{xu2022vitpose} and YOLO11x~\cite{JocherUltralyticsYOLO2023} to generate training pseudo-labels, \NuestroMetodo{} is completely self-contained and detector-free at inference. Fig.~\ref{fig:modeldiag} provides an overview of the model, which is described in detail below.

\subsection{Video Transformer and heatmap processing}
Consider a video segment, or clip, with dimensions $T\times C\times H\times W$ where $T$ is the number of frames and $C, H, W$ are the channels, height, and width of each frame, respectively.
To process this clip with a video transformer~\cite{wang2023videomaev2}, we use the joint space-time cube embedding~\cite{arnab2021vivit}. This technique samples non-overlapping cubes from the input video clip, which are then fed into the embedding layer. This method segments a video sequence
by using cubes of dimension $ {\rm I\!R}^{2 \times C \times 16 \times 16}$, resulting in $V\in {\rm I\!R}^{t\times C\times h\times w}$, where $t= \frac{T}{2}, h=\frac{H}{16}, w=\frac{W}{16}$.
We then project $V$ to a token of dimension $D$ using a linear embedding layer, resulting in an input tensor of shape $X_{vis} \in {\rm I\!R}^{N_{vis} \times D}$, where $N_{vis}=t\cdot{}h\cdot{}w$. Next, we apply a positional embedding to each token, and a learnable class token, $X_{cls} \in {\rm I\!R}^{1 \times D}$, is concatenated into the sequence.
For the computation of heatmaps, our model incorporates $N_{hm} = hm_{res}\cdot{}hm_{res}$
learnable tokens in the input sequence defined as $X_{hm} \in {\rm I\!R}^{N_{hm}\times D}$, $hm_{res}$ defines the heatmap resolution and the number of tokens that represent the heatmap. The complete sequence of tokens, including the class, heatmap, and video tokens $X=(X_{cls}, X_{hm}, X_{vis})\in {\rm I\!R}^{N\times D}$ where $N=1+N_{hm}+N_{vis}$, is then processed through a standard ViT architecture. The processed class token $X_{cls}$ is used in a multilayer perceptron (MLP) named \emph{Class head} to decode the driver distraction class, while the $X_{hm}$ heatmap tokens are passed through the \emph{Heatmap head} to decode the driver's pose and the interacting object. See Fig.~\ref{fig:modeldiag} for an overview of the inputs and outputs of the model.

\subsection{Object and body landmarks detection}

\begin{figure}
    \centering  
    \includegraphics[width=\columnwidth]{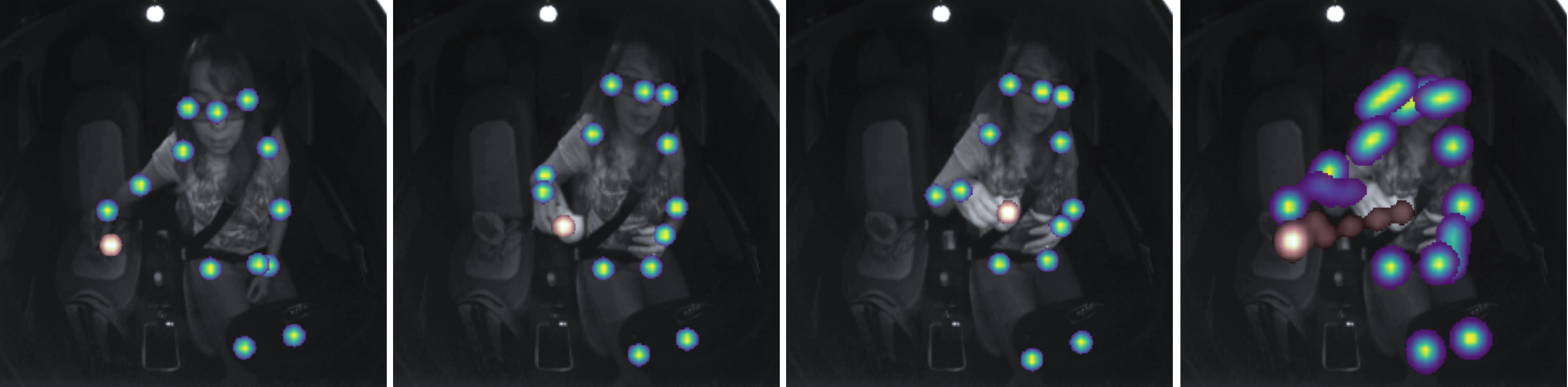}
    \caption{Example of motion heatmaps. The first three images show different frames of a sample clip. The last image shows the result after averaging all heatmaps in the clip. This motion heatmap, one for each item of interest, is used to train our models.}  
    \label{fig:hm_gt}  
\end{figure} 
We approach this task using temporal heatmaps to estimate time-varying body pose and object location. These heatmaps are derived from a common set of learnable tokens, with a fixed number of tokens, which are independent of the prediction tasks for which they are used. This allows extending the classical human pose regression task to predict the landmark location of a varying number of actors and objects without changing the transformer architecture.

Heatmap prediction starts with the introduction of additional tokens to the network, $X_{hm}$, which are randomly initialized at the start of training.
After being passed through the transformer, these tokens are processed by a lightweight decoder to transform them into heatmaps. 
The decoder's architecture consists of two deconvolution layers followed by a convolution layer with a 1 × 1 kernel and with output channels equal to the number of items of interest $L$~\cite{xu2022vitpose}. The output of this decoder is then directly compared with the ground-truth heatmaps by measuring the mean-squared error.

Although the $X_{hm}$ tokens can predict heatmaps for individual frames in a video clip, we can adapt them to capture the entire sequence of movements by modifying the ground truth labels. We create time-aware heatmaps by averaging the frame heatmaps from the ground-truth labels across the entire video clip. This results in a ground truth motion heatmap where body joints and object movement within the clip are visible (see Fig.~\ref{fig:hm_gt}). 

For our specific task, we use these motion heatmaps to embed information about body poses and object location in clips of $T$ frames. For the object localization task, we represent the object with which the actor interacts and its movement throughout the scene. This is done by indicating the center of the object using a Gaussian distribution as ground truth heatmap in every frame of the input clip and then creating the motion heatmap by averaging them. This additional information ensures that the network is focused on
the interacting objects in the scene.

\subsection{Training losses}

We train both the distraction classification and localization heatmap estimation tasks simultaneously. The training loss is defined as follows:  
  
\begin{equation}  
\mathcal{L}_{\text{CE}} = -\frac{1}{N_{batch}} \sum_{i=1}^{N_{batch}} \sum_{c=1}^{C} y_{i,c} \log(\hat{y}_{i,c})  
\label{eq:cross_entropy}  
\end{equation}  
  
\begin{equation}  
\mathcal{L}_{\text{MSE}} = \frac{1}{N_{batch}} \sum_{i=1}^{N_{batch}} ||L_i - \hat{L}_i||_F^2  
\label{eq:mse}  
\end{equation}  
  
\begin{equation}  
\mathcal{L} = \mathcal{L}_{\text{CE}} + \log \left( \mathcal{L}_{\text{MSE}} \right)  
\label{eq:overall_loss}  
\end{equation}  
  
\(\mathcal{L}_{\text{CE}}\) is the cross-entropy loss for the classification task as defined in \eqref{eq:cross_entropy}, and \(\mathcal{L}_{\text{MSE}}\) is the mean-squared error loss for the heatmap prediction task as defined in \eqref{eq:mse}, where $||\cdot||_F$ is the Frobenius norm. This directly measures the element-wise error between the predicted and ground-truth motion heatmap. Here, \(y_{i,c}\) and \(\hat{y}_{i,c}\) are the true and predicted driver action class probabilities for driver action \(c\) of sample \(i\), respectively, and \(L_i\) and \(\hat{L}_i\) are the true and predicted heatmap values, respectively. The overall loss function \(\mathcal{L}\) is given by \eqref{eq:overall_loss}, we have log scaled \(\mathcal{L}_{\text{MSE}}\) for better multi-task training. Further discussion of task balancing is given in the ablation study.

\subsection{Pose-and-object-GUIded token SElection module} 

Processing videos using joint space-time cube embeddings is computationally expensive, making it impractical for environments with limited computing power, such as onboard vehicle systems. In-cabin videos often contain repetitive information over time and areas that are devoid of relevant data for driver action recognition. To address this, we introduce an enhanced set of features for the token selection module, which reduces computation while retaining crucial task-relevant information in the car driving context.

This module operates through a two-step token selection process. Initially, in token pruning, the most relevant vision tokens ($X_{vis}$) are selected based on their attention to the heatmap and the class tokens. The number of tokens selected in this first step is determined by the \textit{token keep rate} $\rho$, which results in $N_{prune} = N_{vis} \cdot \rho$. In the second step (token merging), the module matches and merges the discarded tokens from the previous step based on their similarity, retaining the most similar tokens $N_{merge} = N_{disc} \cdot \lambda$ to minimize information loss, with $N_{disc}=N_{vis}-N_{prune}$. After the token selection module, the number of tokens in $X_{vis}$ is $N_{vis}=N_{prune} + N_{merge}$. Both $\rho$ and $\lambda$ are predefined \textit{keep rates} within the range (0, 1] and directly control the computational cost of the network. The token selection module is integrated into the transformer network architecture at specific intervals, effectively reducing computational load while maintaining task-relevant information. 

Our approach, \NuestroMetodo{}, further enhances token selection by incorporating a richer set of heatmap features that include interacting objects, making the module more effective for the driver distraction detection task. 
Specifically, we take advantage of the flexibility of our heatmap representation $X_{hm}$ to add the task of localizing the interacting object. This new task is treated as an additional item of interest as the $L$-th ground-truth label. This enhancement does not require any changes to the underlying architecture, but significantly improves token selection and the training process, perfectly adapting our model to the task of driver distraction detection. This enhancement is particularly notable at low \textit{keep rates}, where the inclusion of objects results in a more refined token selection. This improvement is illustrated in Fig.~\ref{fig:token_sel} of our experiments, which provides a comparison of the tokens selected with and without the inclusion of object-augmented heatmaps.

\section{Experiments}
\label{sec:exps}

In this section, we evaluate our multi-task video transformer. In all the experiments \emph{HM} stands for heatmaps computed for: \emph{P} single human pose (body joints location) or \emph{OB} object location.
Within each experiment, the results of the \first{first}, \second{second}, and \third{third} ranked are shown, respectively, in bold, underlined, or double underlined.

\subsection{Datasets} 

\begin{figure}[t]
    \centering  
    \includegraphics[width=\columnwidth]{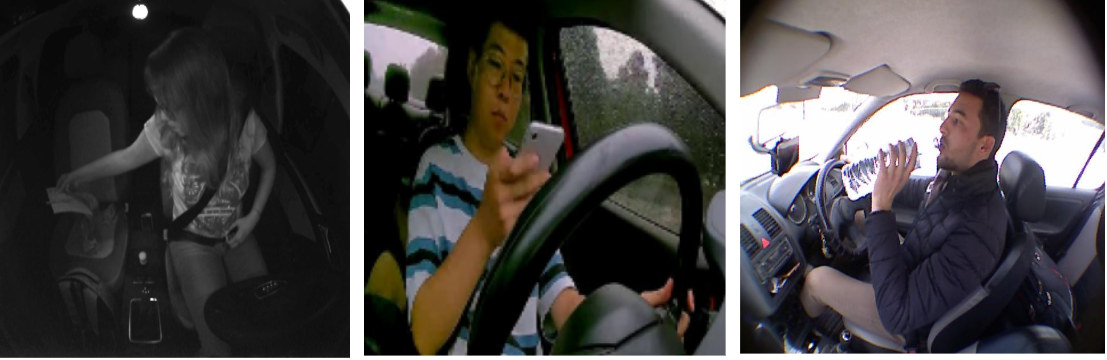}
    \caption{Sample images from each dataset. We use the same view in our experiments. Left: Drive\&Act, fetching\_an\_object. Middle: 100-Driver, Talking right hand. Right: 3MDAD, Drinking using right hand.}  
    \label{fig:example_classes}  
\end{figure}

We evaluated our method using three datasets of driving action recognition, each with slight variations in context, such as autonomous versus manual driving, camera views, modalities, and distraction classes. Figure~\ref{fig:example_classes} shows examples from each dataset and the camera view used in the experiments.

Drive\&Act~\cite{martin2019driveact} is a multi-modal driver action recognition dataset containing 12 hours of driving over 15 different drivers and different views. We use fine-grained labels that consist of 34 unique activities that a driver might engage in a fully autonomous vehicle, such as eating, working on a laptop, fetching or placing an object, among others. Each labeled clip corresponds to a segment of 3 seconds or less.
The dataset is divided into three predefined splits/folds for training and evaluation with no driver overlap. We report the average of the results on the three test sets. An issue with this dataset is the high imbalance between the classes. As such, there is a need to make a distinction between the metrics reported. For consistency with previous works, we report both the top-1 accuracy (micro accuracy) and the average per-class accuracy (macro accuracy). However, we will only use the macro accuracy in our discussion, since it takes into account the class imbalance and gives a better understanding of the model's overall performance. We use only the front-top (\textit{inner\_mirror}) view taken from a NIR camera for comparison with previous methods. 

100-Driver~\cite{wang2023100driver} is a large-scale distracted driver dataset. It provides more than 470K images taken by 4 cameras observing 100 drivers over 79 hours from 5 vehicles. It contains different types of variations that closely meet the real-world
applications, including changes in vehicle, person, camera view, lighting, and modality. In total, it has 22 classes, one of which is normal driving. In our experiments, we use the predefined traditional evaluation setting where train, validation, and test splits use different drivers. We use the Day recordings with Camera 3, which is located in the dashboard. Each label indicates the peak frame of the action, and additionally, we sample frames around this peak frame.

3MDAD~\cite{jegham20203mdad} is another dataset that offers synchronized multimodal and multi-view data. It includes “safe driving” and 15 common distraction activities performed by 50 drivers in the daytime, with a total of 444,104 frames captured by 5 different cameras. Although it does not provide a predefined train / test split, we use the methodology of previous works, we employ k-fold cross-validation with k = 5. In each fold, the dataset is divided into a 40/10 ratio for the train/test split, ensuring that the drivers appearing in the test subset are not present in the train. All experiments are done on the RGB modality using the side camera.

\subsection{Implementation details} 

In our experiments, we include \NuestroMetodo{} into two ViT-based backbones. The first architecture is VideoMAEv2~\cite{wang2023videomaev2}, a ViT-based model with pre-trained weights distilled from ViT-giant (\textit{vit\_b\_k710\_dl\_from\_giant}). It processes input clips of size $T=16$. The second architecture is InternVideo2~\cite{wang2024internvideo2}, a model bigger than VideoMAEv2, similarly pre-trained, and with slight changes in the layers used. It processes input clips of size $T=8$. We utilize the weights InternVideo2-B/14 distilled.

We use the AdamW~\cite{loshchilov2018decoupled} optimizer with a Cosine Annealing learning rate scheduler~\cite{loshchilov2017sgdr} with $T\_max=30$. To train both tasks simultaneously, we minimize a joint objective function defined as:
\begin{equation*}
    \mathcal{L}_{total}^{(t)} = \lambda_{cls}^{(t)} \mathcal{L}_{CE}(y, \hat{y}) + \lambda_{hm}^{(t)} \log\left( \mathcal{L}_{MSE}(H, \hat{H}) \right)
    \label{eq:total_loss}
\end{equation*}
where $\mathcal{L}_{CE}$ is the Cross-Entropy loss and $\mathcal{L}_{MSE}$ is the Mean Squared Error for heatmap prediction. We apply a logarithmic transformation to the MSE term to normalize its magnitude relative to the classification loss. To dynamically balance the gradients between these objectives, we employ Nash-MTL~\cite{navon2022multi}, which computes the optimal coefficients $\lambda_{cls}^{(t)}$ and $\lambda_{hm}^{(t)}$ every 20 training steps $t$, initialized as $\lambda_{cls}^{(0)} = \lambda_{hm}^{(0)} = 1.0$.

Data augmentation includes Cutmix~\cite{yun2019cutmix}, Mixup~\cite{zhang2018mixup} and RandAug~\cite{cubuk2020randaugment}.
During training, we randomly sample frames from the input clip and randomly crop part of the frames, maintaining the full height and aspect ratio, and resize it to 224×224 pixels. During inference, we crop the central part of the frame, maintaining the full height and aspect ratio, and resize it to 224×224 pixels. For each labeled clip, we sample frames at equal time intervals and use these frames as input to the network, where the first and last sampled frame correspond to the first and last instant of the labeled clip. For each dataset, we estimated the human pose for each frame using ViTPose~\cite{xu2022vitpose} and the object localization using YOLO11x~\cite{JocherUltralyticsYOLO2023}, followed by manual refinement of the detected objects. The estimated body joint and object localizations are used for generating our ground truth heatmaps in all experiments. During refinement, we cross-referenced the object detections with the existing action class labels to verify that the localized objects corresponded to the interacting items specified by the class (e.g., a cellphone for 'Talking phone' or a bottle for 'Drinking'). It is crucial to note that these external tools are used exclusively for creating the training data and are not required during inference.  For the total number of heatmaps $L$, Drive\&Act uses 14, 100-Driver 11 and 3MDAD 14. Detailed hyperparameters for the experiments are available in the code repository.

Finally, to assess the feasibility of onboard deployment, we conduct efficiency benchmarks on an NVIDIA Jetson Orin NX (16GB) embedded computer. This standardized inference suite is utilized in all experiments in which speed is measured. The device is configured in MAXN power mode to allow maximum power consumption and performance, with the fan speed maximized to prevent thermal throttling during testing. Inference is executed with FP16 and batch size of 1, simulating an online processing pipeline. To ensure precise and reproducible measurements, we calculate the average over 1,000 inference runs, excluding the initial model warm-up phase and data loading overhead. Time measurements are synchronized using PyTorch's CUDA events (\texttt{torch.cuda.Event} and \texttt{torch.cuda.synchronize}) to accurately capture the asynchronous GPU execution time.

\subsection{Ablation study}
\label{sec:ablation}

For the ablation experiments in this section (see Tables~\ref{tab:ablation},\ref{tab:abl_model_hm},\ref{tab:driveacthmtabl}), we use the Drive\&Act dataset fold 0 for training and testing. Our baseline result is obtained by fine-tuning a state-of-the-art video transformer, VideoMAEv2~\cite{wang2023videomaev2} pre-trained in Kinetics~\cite{carreira2017i3d}.  The accuracy for the baseline for the fold 0 in Drive\&Act is 69.08\%.

\begin{table}
  \caption{Ablation study. Test results on fold 0 of Drive\&Act dataset using different model configurations.}
  \centering
  \renewcommand{\arraystretch}{1.2}
  \begin{tabular}{l|c|c}
    \toprule
     \multirow{2}{*}{Method} & Macro Acc. & GFlops \\
     & ($\uparrow$) & ($\downarrow$)  \\
    \hline
    VideoMAEv2-base & 69.08 & \third{360} \\
    +HM(P) & 70.15 & \second{381} \\
    +HM(P+OB) & \second{72.16} & \second{381}\\
    \hline
    PO-GUISE & \third{71.05} & \first{251} \\
    \hline
    \NuestroMetodo{} & \first{72.62} & \first{251} \\ 
  \bottomrule
  \end{tabular}
  \label{tab:ablation}
\end{table}

\begin{table}
    \caption{Ablation on different heatmap resolutions. Test results on Fold 0 of Drive\&Act dataset.}
    \centering
    \renewcommand{\arraystretch}{1.2}
    \begin{tabular}{c|c|c}
         \toprule
         Method & Macro Acc. & GFlops\\
         & ($\uparrow$) & ($\downarrow$) \\
         \hline
         \NuestroMetodo{} (2x2) & 69.69 & \first{236} \\
         \hline
         \NuestroMetodo{} (4x4) & \third{69.78}& \second{239} \\
         \hline
         \NuestroMetodo{} (8x8) & \first{72.62} & \third{251} \\
         \hline
         \NuestroMetodo{}(16x16) & \second{70.71} & 269 \\
         \bottomrule
         
    \end{tabular}
    \label{tab:abl_model_hm}
\end{table}

\begin{figure}[ht]  
    \centering  
    \includegraphics[width=\linewidth]{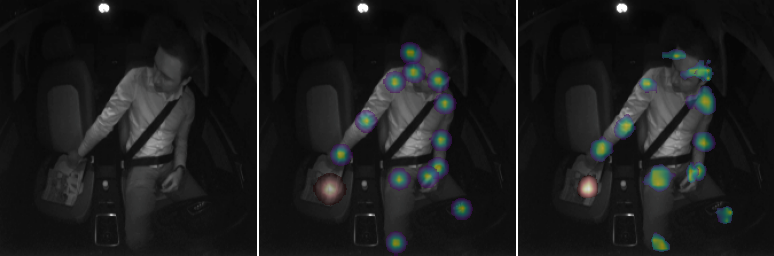}
    \includegraphics[width=\linewidth]{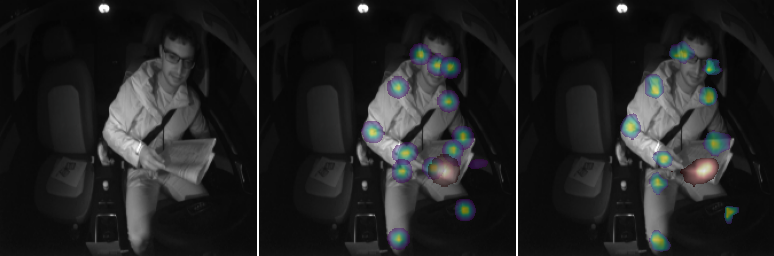}
    \caption{Sample heatmaps used in our method. The first column shows the raw image, the second the ground truth heatmaps used in training, and the third column shows the prediction of our network. We have used a different color for the object location heatmap.}  
    \label{fig:example_hm}  
\end{figure} 

\begin{table}
  \caption{Heatmap estimation results in Drive\&Act measured by Average Intersection over Union (IoU). We compare \NuestroMetodo{} without task scaling against models with different task scaling techniques.
  }
  \label{tab:driveacthmtabl}
  \centering
  \renewcommand{\arraystretch}{1.2}
  \begin{tabular}{@{}l|c@{}}
    \toprule
    Method & Avg. IoU ($\uparrow$)\\
    \hline
    Mean driver position & 0.184\\
    \hline
    Base & \third{0.195} \\
    + log scaling & \second{0.291} \\
    + Nash-MTL\cite{navon2022multi} & \first{0.342} \\
  \bottomrule
  \end{tabular}
\end{table}

\subsubsection{Comparison with baseline.}
First, we test the baseline plus semantic information in the form of a human pose estimation task, see baseline+HM(P) in Table~\ref{tab:ablation}. On average, it increases the accuracy of all actions by 1.07 points in Drive\&Act. The addition of objects, see baseline+HM(P+OB), increases the baseline by 3.08\%. The pose and object information provides a significant improvement in the accuracy of some actions. A small drawback is the increased computational cost of 5\% more GFLOPS, due to the additional tokens that need to be processed for the heatmap estimation.

\begin{figure}
\centering
\includegraphics[width=\columnwidth]{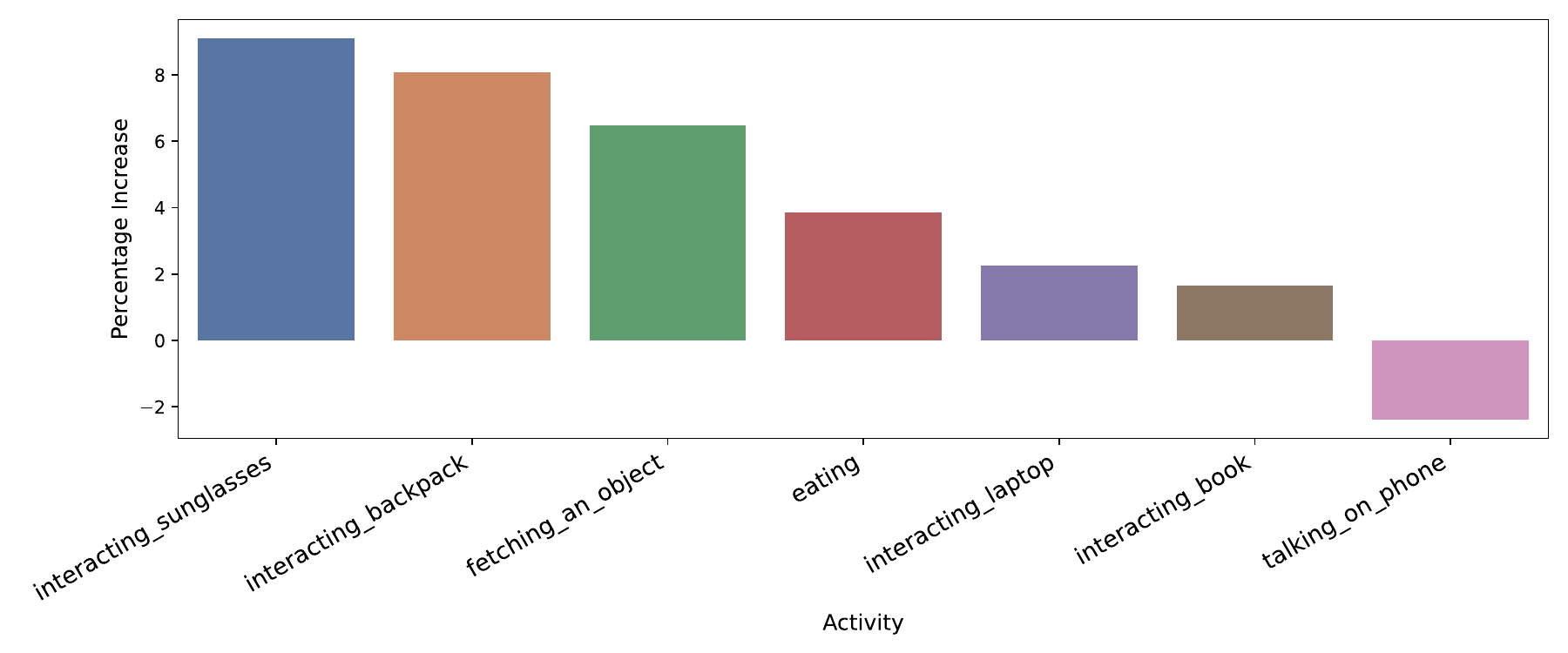}%
\caption{Sample classes where \NuestroMetodo{} obtains a difference in performance when compared to guiding token selection without object heatmap. We have merged some classes for an easier comparison.}
\label{fig:obj_inc}
\end{figure}

The last set of experiments assesses the influence of using our enhanced heatmap estimation for token selection. The PO-GUISE module improves the baseline by 1. 97\% while reducing GFlops by 30\%. Our method \NuestroMetodo{}, which adds the guidance of object location in addition to pose, increases the accuracy by 3.54\%, while matching the same number of GFlops. In fact, some actions that involve object interaction, such as sunglasses, backpack, food, or fetching objects, get the greatest improvement, around 4.14\%. Fig.~\ref{fig:obj_inc} shows the classes in which we observed a difference in performance using \NuestroMetodo{}. There is an increase in accuracy across many object-related classes, with a minor decrease in only one class. This decrease is due to the challenge of detecting a cell phone when a person is talking, given the angle of view of the camera used. For this analysis, we have merged some time-related classes for easier comparison. For example, the class `eating' includes `preparing\_food' and `eating'.

Finally, regarding the preservation of spatiotemporal information, the superior accuracy of \NuestroMetodo{} over the unpruned baselines (VideoMAEv2-base and +HM(P+OB)) suggests that our guided selection mechanism functions as an effective semantic filter. By explicitly prioritizing tokens with high attention to the pose and object heatmaps, the model effectively eliminates visual noise and temporal redundancy without losing the critical features required to distinguish fine-grained distraction behaviors.

\subsubsection{Heatmap estimation} 

In this section, we validate the performance of the heatmap estimation of our model and the techniques used to balance both tasks, driver distraction recognition, and heatmap localization.

An important parameter in our model, $hm_{res}$, dictates the grid size of heatmap tokens, i.e., the number of tokens that represent an image of size $224\times224$ pixels. A higher $hm_{res}$ means that the image is represented with more tokens, allowing for more fine-grained details. However, this comes at the cost of increased computational resources because more tokens need to be processed by the transformer. In Table~\ref{tab:abl_model_hm}, we present the results of our ablation study in different $hm_{res}$. The results indicate that an $hm_{res}$ of 8$\times{}$8 is optimal for this setting. Although an increase in grid size is usually beneficial in general human-pose estimation tasks~\cite{xu2022vitpose}, we have found that for this task, where the subjects of interest, the driver and the objects, have a fixed scale, an increase of $hm_{res}$ is not advantageous.

Table~\ref{tab:driveacthmtabl} presents the prediction results measured by the Average Intersection over Union (Avg. IoU) between the predicted heatmaps and the ground truth under different settings. The Avg. IoU is computed by thresholding the heatmaps (threshold=0.1), converting them into binary masks and calculating the IoU, which directly assesses the overlap between the predicted and ground-truth heatmaps, as well as the quality of the heatmap estimation. We calculated the Avg. IoU on Fold 0 of the Drive\&Act dataset using \NuestroMetodo{}. To better assess the impact of driver movement, classes where no driver is present or classified as ``sitting still" were excluded from the analysis.
To contextualize the IoU values, we measured the mean driver position IoU, which represents the IoU achieved if the heatmap prediction defaults to the mean driver position in the data set. In the ``Base" setting, both classification and estimation tasks are trained simultaneously without scaling or modifying the loss functions. This approach achieves an IoU of 0.195, indicating that while the predictions slightly improve over the mean driver position, they fail to effectively capture the driver’s movements. Applying logarithmic scaling to the heatmap loss leads to an improvement in IoU of 0.291. Furthermore, incorporating Nash-MTL~\cite{navon2022multi}, a technique to dynamically balance the gradients resulting from the losses of both tasks, significantly enhances performance, achieving an IoU of 0.342. At this stage, the predictions accurately determine the center of each heatmap’s location, but the IoU remains limited due to inaccuracies in predicting the heatmap boundaries and intensities. Figure~\ref{fig:example_hm} illustrates sample predictions alongside their corresponding ground truth. We obtain an Avg. IoU of 0.348 and 0.403 for each sample, respectively.

\subsection{Efficiency analysis} 
\begin{figure}
\centering
\includegraphics[width=0.6\columnwidth]{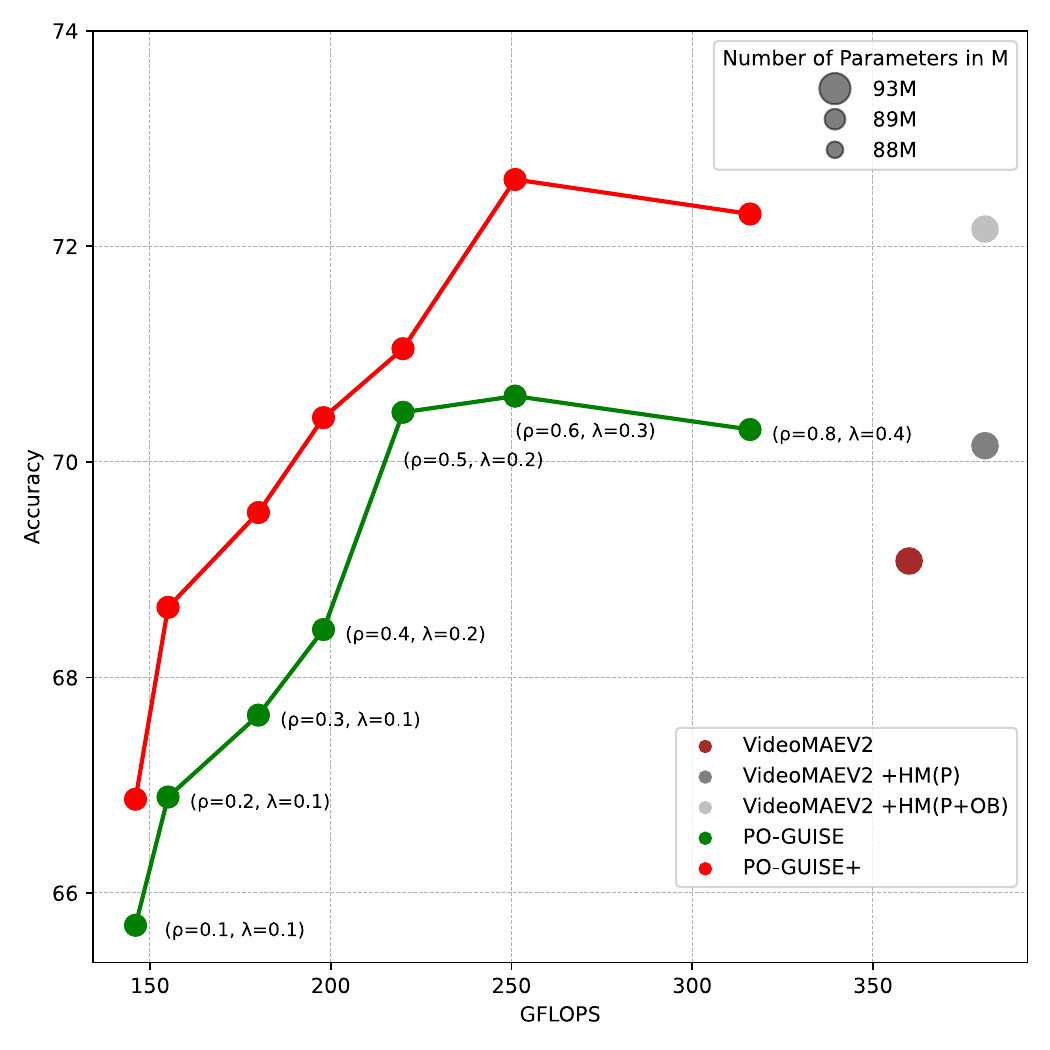}%
\caption{Comparison between GFLOPS and accuracy for different configurations and top methods from SOTA in Drive\&Act. Circle size represents the number of parameters. For each experiment, both \NuestroMetodo{} and PO-GUISE are matched in terms of GFLOPs and token keep rate. }
\label{fig:flop_comp}
\end{figure}

\begin{figure}
\centering
\includegraphics[width=0.6\columnwidth]{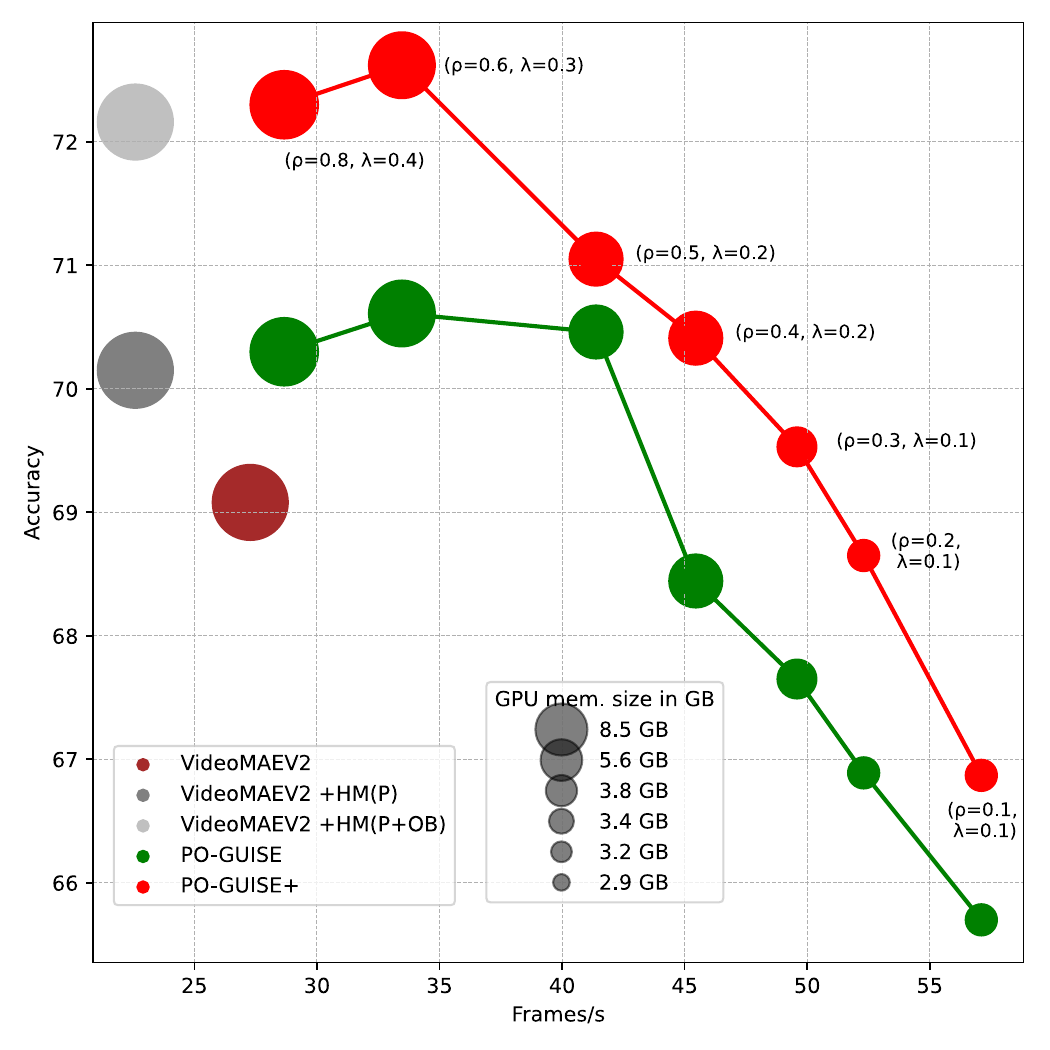}%
\caption{Comparison between clips per second and accuracy for different configurations and top methods from SOTA in Drive\&Act, evaluated in a Jetson Orin NX. Circle size represents GPU memory used at inference.}
\label{fig:clip_s_comp}
\end{figure}

In this experiment, we explore the trade-off between accuracy and computational cost incurred by different token selection rates in PO-GUISE and our improved \NuestroMetodo{} with VideoMAEv2 as the backbone model. We use the Drive\&Act dataset fold 0 for training and testing.
In Figures~\ref{fig:flop_comp} and~\ref{fig:clip_s_comp} we show, respectively, the accuracy vs GFLOPS and accuracy vs frames/s curves for different values of $\rho$ and $\lambda$, $\rho\in\{0.1,0.2,0.3, 0.4,0.5, 0.6,0.8\}$ and $\lambda\in\{0.1,0.1,0.1,0.2,0.2,0.3,0.4\}$. These configurations were selected to demonstrate different computational budgets. The model accuracy-efficiency trade-off is more favorable when $\lambda$ is substantially lower than $\rho$. This ensures that token pruning is the primary selection mechanism, while merging is used more sparingly to recover critical information from a subset of the discarded tokens.

The curve associated with \NuestroMetodo{} is always above for different proportions of selected tokens. Even at a very low keep rate, using 146 GFlops, the accuracy of our method is 66.87\%, a 1.17\% increase over PO-GUISE. This highlights the importance of taking into account the interaction with objects in guiding the token selection process for driving action recognition. Interestingly, we observe a slight decrease in performance at the highest computational budget (316 GFLOPS), where very few tokens are being pruned. This result suggests that our token selection mechanism may provide a regularizing effect. By forcing the model to make predictions from a reduced set of tokens, it may prevent overfitting to unrelated features in the training data. As the token keep rate approaches 100\%, this regularization benefit is diminished, leading to a marginal decrease in generalization performance.

\begin{figure}
\centering
\includegraphics[width=0.6\columnwidth]{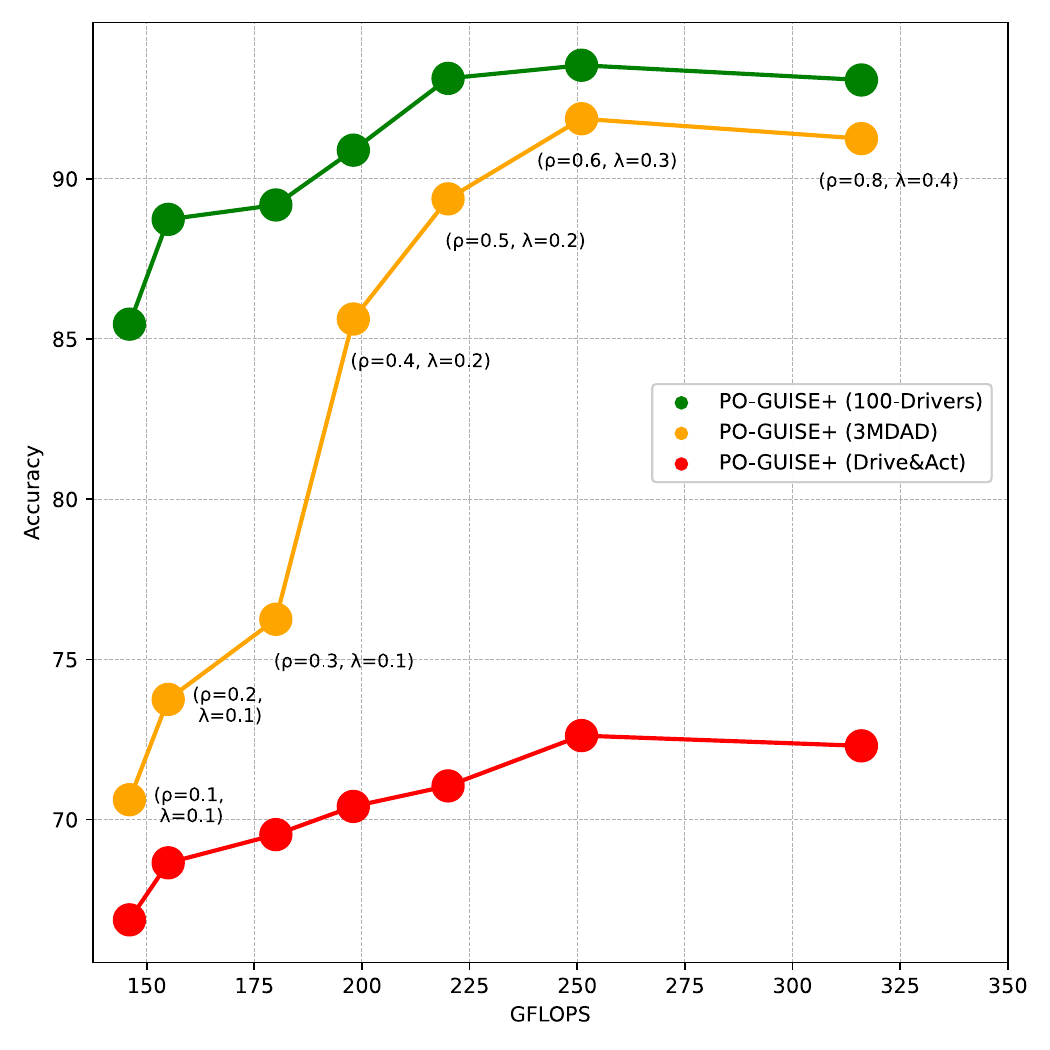}%
\caption{Comparison between GFLOPS and accuracy for different keep rate configurations across different datasets.}
\label{fig:kp_datasets}
\end{figure}

To validate the generalizability of this trade-off, we extended the analysis to the 100-Driver and 3MDAD datasets (see Fig.~\ref{fig:kp_datasets}). While the regularization pattern persists, 3MDAD shows a steeper performance drop at lower budgets. We attribute this sensitivity to the significantly smaller size of the 3MDAD dataset. Furthermore, this comparison reveals that determining the optimal keep rate is not only a function of the desired GFLOPs, it depends on the dataset size, complexity of the scenario, and the camera position. For instance, the robust performance on 100-Driver suggests that larger datasets or favorable camera angles allow for aggressive pruning, whereas restricted views or smaller datasets (like the side-view in 3MDAD) require higher keep rates to capture the necessary visual context for accurate classification.

We have also conducted experiments on a Jetson Orin NX (16GB) to evaluate real-world performance, as shown in Fig.~\ref{fig:clip_s_comp}. The baseline models utilize VideoMAEv2, either without or with heatmaps (+HM(P) and +HM(P+OB)), and process clips of 16 frames. These models suffer from high computational costs in terms of memory usage and inference speed. Specifically, VideoMAEv2 with heatmaps (+HM(P) and +HM(P+OB)) uses 5.6GB of memory, achieving 70.15\% and 72.16\% accuracy for the variants that use only the driver pose (+HM(P)) and the driver pose with objects (+HM(P+OB)), respectively. 

Using \NuestroMetodo{}, our best performing model obtains an accuracy of 72. 62\%, with an inference speed of 33 frames per second, using only 3.8GB of memory. This gain in performance over the baseline is especially important in the Jetson architecture, where the GPU and CPU share the same unified memory, meaning that a lower model memory requirement leaves more space for other secondary CPU tasks. Additionally, our smallest model, which uses 2.9GB of memory and achieves an inference speed of 57 frames per second. \NuestroMetodo{} processes clips of 16 frames, meaning that it could process 3.56 clips every second. This is feasible for implementation on lower-end Jetson models while still having better accuracy than previous works.

As another point of comparison, we have evaluated a lightweight CNN, an I3D~\cite{carreira2017i3d} model, which operates at 33 GFLOPS and uses 0.53GB of memory.  Although this architecture allows fast inference at 170 FPS, it achieves a significantly lower accuracy of 46.25\%, making it unsuitable for this safety-critical task.

To provide a more direct comparison, we conducted an additional experiment by training PO-GUISE+ with a smaller pretrained ViT-S backbone with aggressive token pruning ($\rho=0.1$, $\lambda=0.1$). This lightweight configuration operates at 51 GFLOPS, a budget comparable to I3D, and requires 0.65GB of memory. It achieves an inference speed of 105FPS.
and an accuracy of 57.42\%. 
This result demonstrates that for a negligible increase in memory usage, our transformer-based approach outperforms the lightweight CNN by 11.17 points in accuracy. 
While these benchmarks reflect architectural efficiency, total system runtime would include platform-specific overheads (e.g., camera drivers, middleware). In optimized pipelines, these typically add 15-25\% latency. For our 33 FPS model ($\sim$30.3 ms inference), this implies an estimated total of $\sim$35-38 ms per frame, demonstrating real-time feasibility.

\subsection{Qualitative results}

\begin{figure}
    \centering  
    \includegraphics[width=0.8\linewidth]{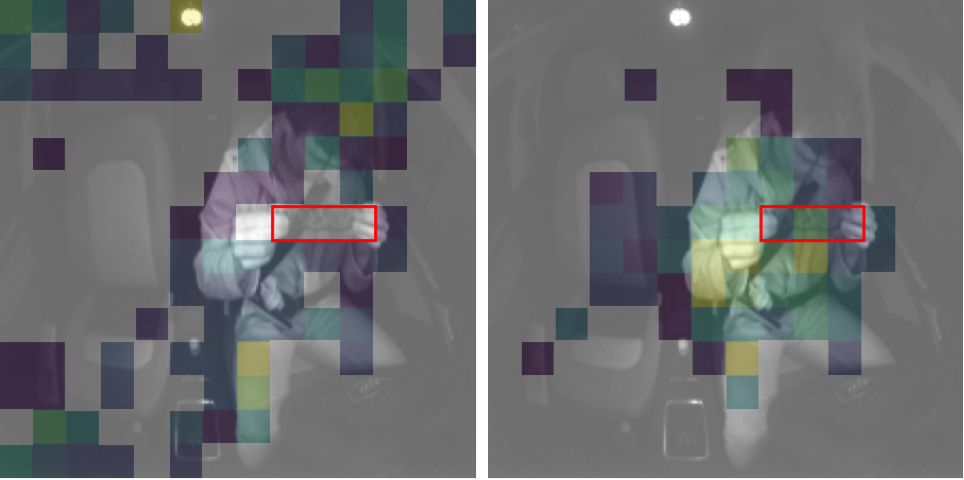}
    \includegraphics[width=0.8\linewidth]{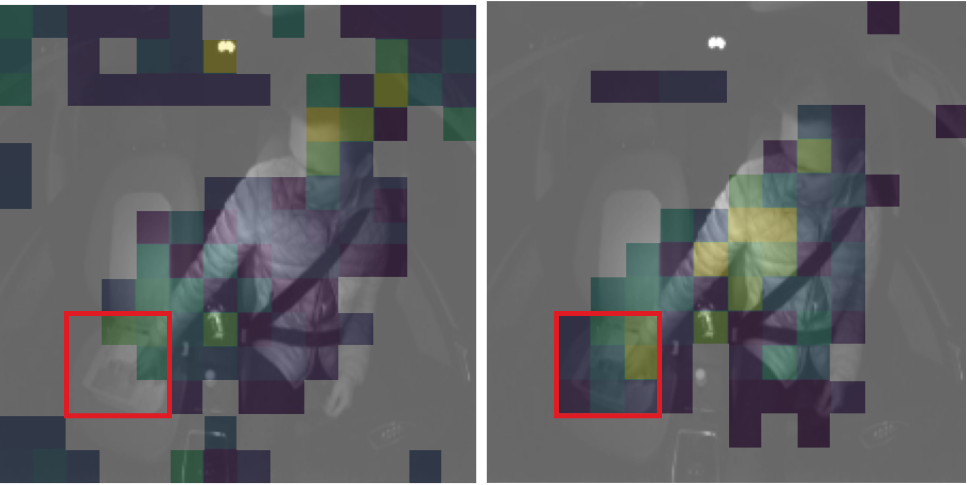}
    \caption{Comparison of token selection in a video clip at a low \textit{keep rate}. The left image shows the tokens selected by PO-GUISE, while the right image shows the tokens selected by \NuestroMetodo{}. Brighter colors indicate that the token was spatially selected more frequently in the video clip, with a grayed-out area indicating that the token was not selected. We have highlighted the interacting object with a red box.}  
    \label{fig:token_sel}  
\end{figure}

In this section, we qualitatively analyze the results of our enhanced heatmap features for token selection and show some example cases for the prediction of our model. Fig.~\ref{fig:token_sel} presents a comparison between PO-GUISE and \NuestroMetodo{}. The figure shows a frame of a video clip partitioned into visual tokens, highlighting those that were selected more frequently in the given clip. Given the same \textit{keep rate} for both models ($\rho=0.2,\lambda=0.1$), \NuestroMetodo{} effectively selects tokens directly related to the driver and the interacting object. In contrast, guiding token selection only by the body pose and class tends to select landmark-related parts of the image but overlooks the interacting object. We believe that this difference in token selection is a key factor that contributes to the superior performance of \NuestroMetodo{} at low \textit{token keep rates}.

\begin{figure}
    \centering  
    \includegraphics[width=0.9\linewidth]{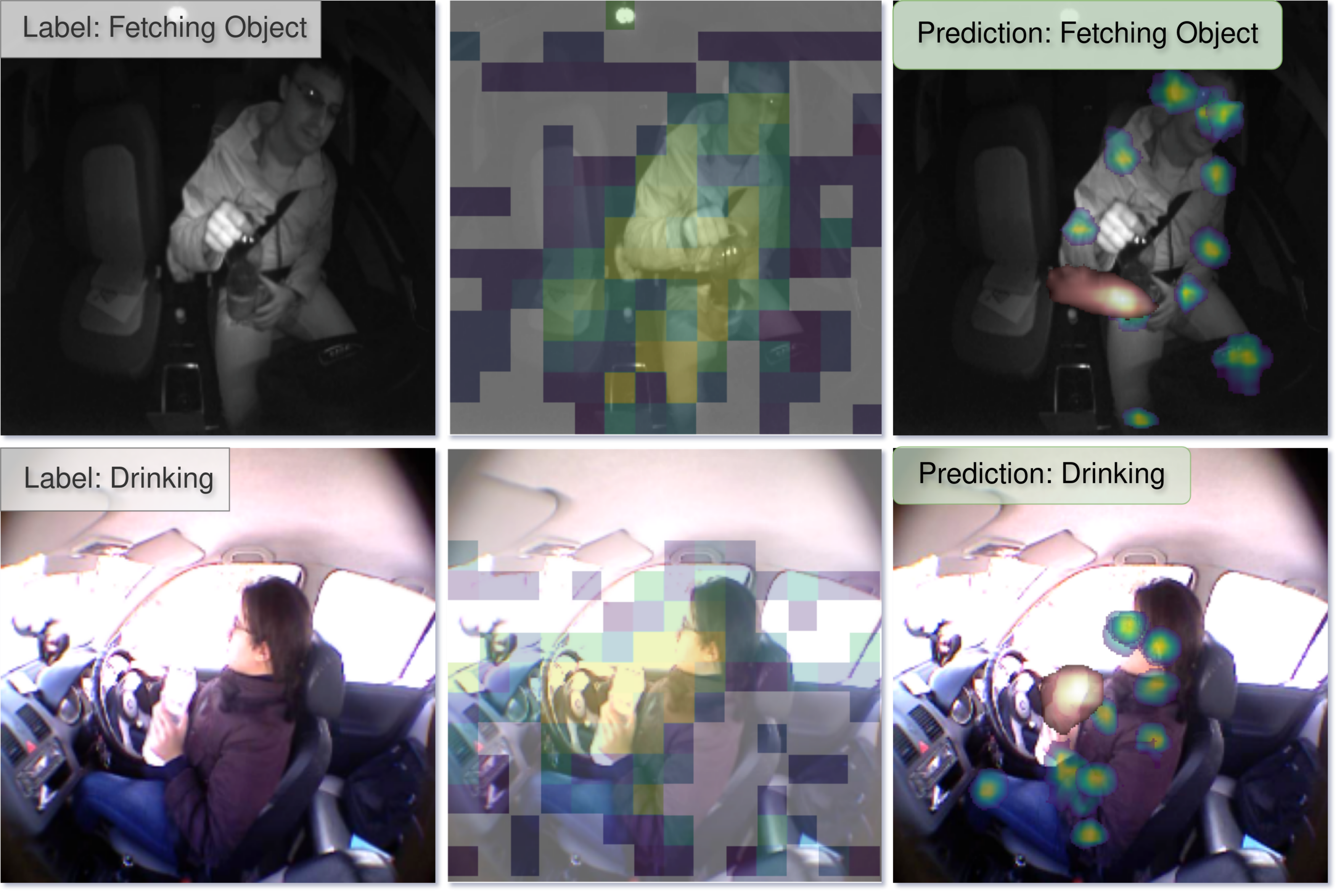}
    \caption{Correct predictions by \NuestroMetodo{} in the datasets Drive\&Act (top) and 3MDAD (bottom), with the classes `fetching\_object' and `Drinking using left hand', respectively. Each row shows a sample frame (left), the tokens selected from the clip (middle), and the predicted heatmaps and label (right).}  
    \label{fig:example_good}  
\end{figure}

\begin{figure}
    \centering  
    \includegraphics[width=0.9\linewidth]{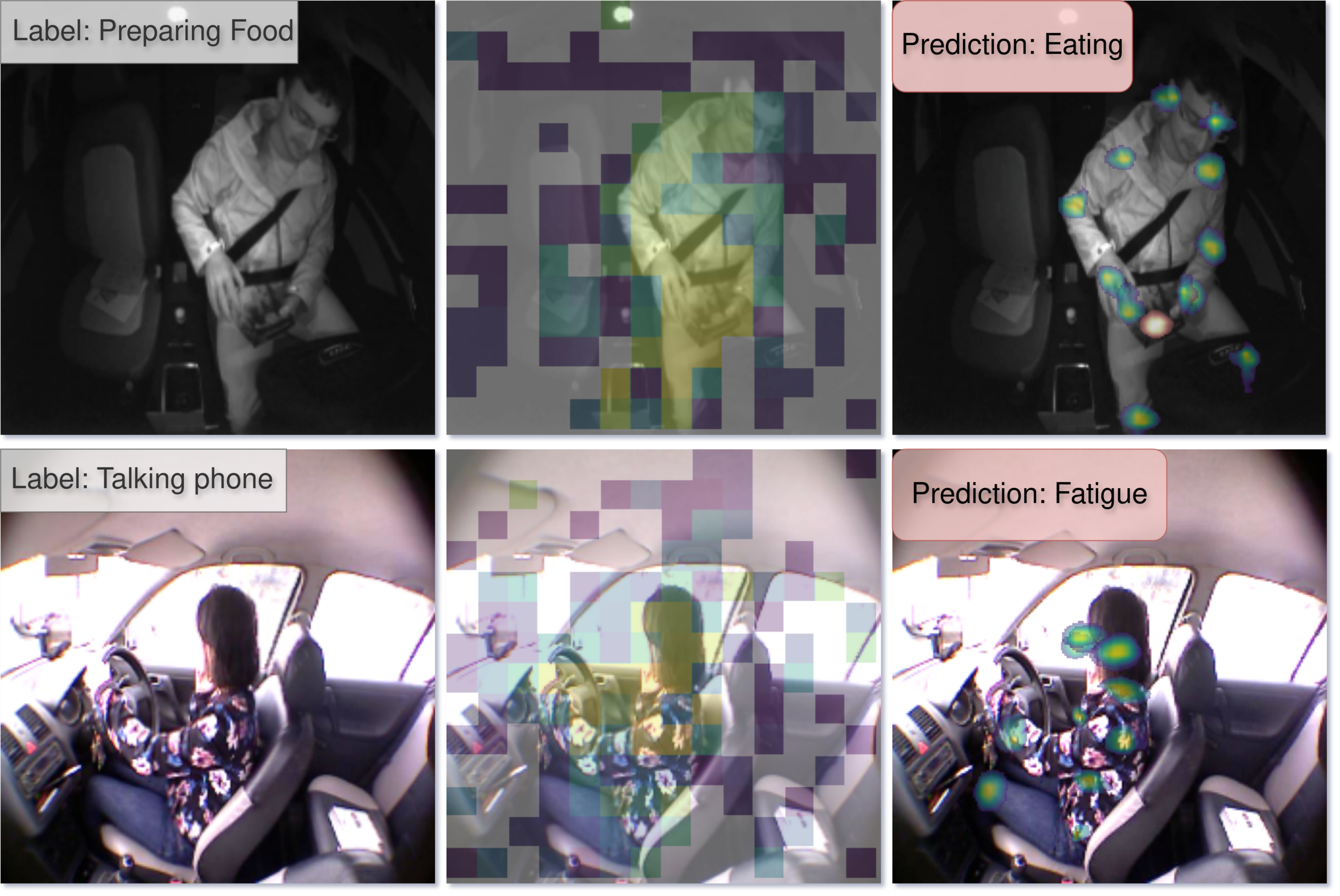}
    \caption{Challenging cases in the Drive\&Act (top) and 3MDAD (bottom) datasets, for the classes `preparing\_food' and `Talking phone using right hand', respectively. Each row presents a sample frame (left), the tokens selected from the clip (middle), and the predicted heatmaps and label (right). Although the token selection and heatmap predictions are accurate, \NuestroMetodo{} fails to classify these correctly due to the limited temporal window and the camera angle.}  
    \label{fig:example_fail}  
\end{figure}

Figures~\ref{fig:example_good} and~\ref{fig:example_fail} show example predictions of \NuestroMetodo{}. In Fig.~\ref{fig:example_good}, the model correctly predicts the class `fetching\_object' and accurately tracks the object throughout the clip, as indicated by the red oval-shaped heatmap corresponding to the object. In contrast, Fig.~\ref{fig:example_fail} presents some failure cases for \NuestroMetodo{}. In the Drive\&Act dataset, most failures, including the one depicted, are due to a lack of temporal context. This means that some distractions are visually similar and cannot be distinguished when observing only a 3-second clip of the video. This is an issue inherent to this task and the way in which this dataset has been structured. Meanwhile, in the 3MDAD dataset, the failure is attributed to the camera angle, which obscures the phone.

\begin{figure}
    \centering  
    \includegraphics[width=\linewidth]{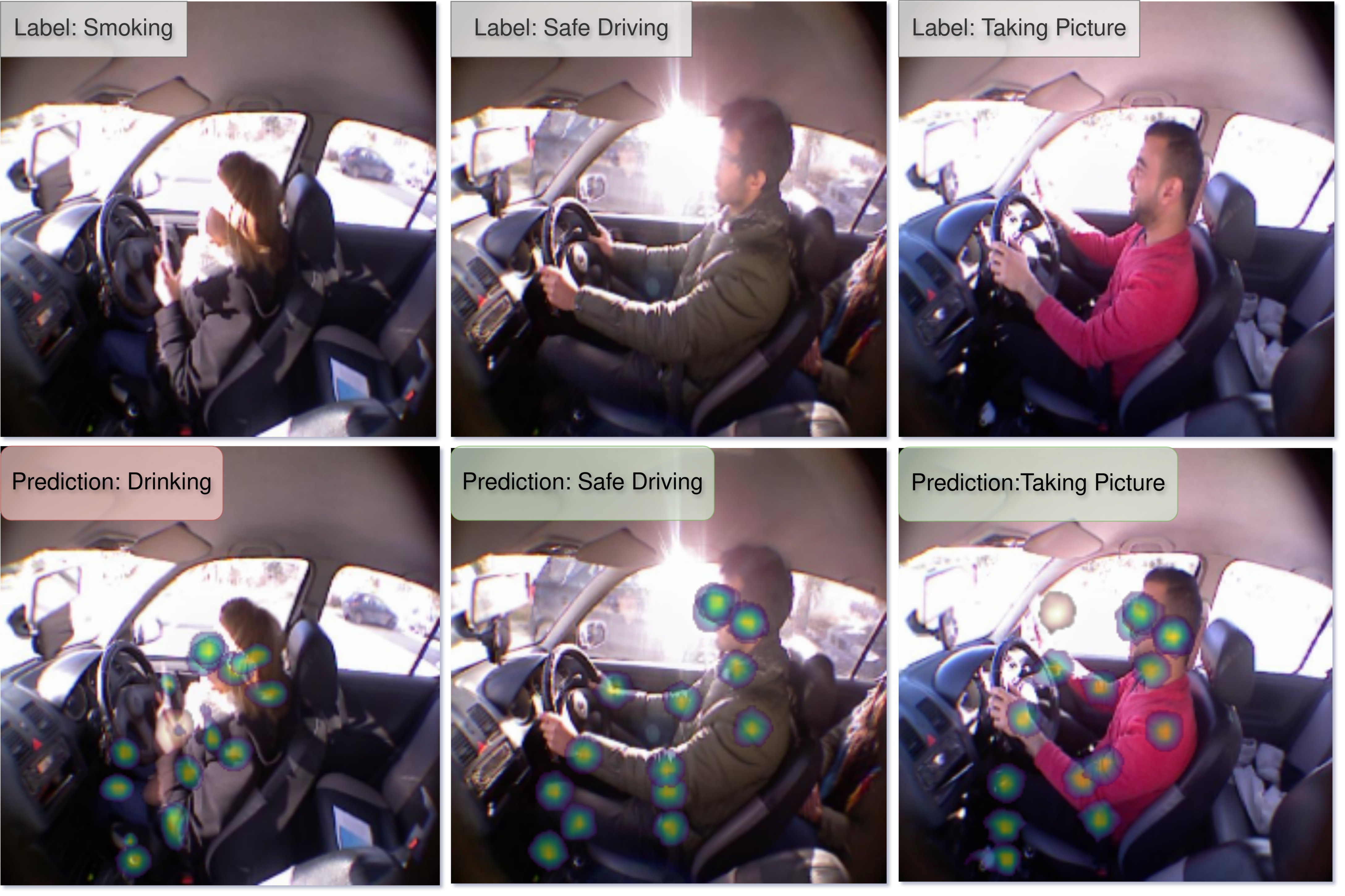}
    \caption{Qualitative results on the 3MDAD dataset under challenging lighting conditions. (Left) An example of a misclassification, where the action was incorrectly identified as 'Drinking'. (Middle) Despite sun glare obscuring parts of the scene, the model correctly classifies the action as 'safe driving'. (Right) The model successfully identifies the driver 'taking a picture'. These examples highlight the model's robustness to common and challenging visual conditions found in real-world driving scenarios.}  
    \label{fig:example_light}  
\end{figure}

Beyond these task-specific challenges, robust real-world deployment requires addressing significant environmental variations, particularly in lighting. Fig.~\ref{fig:example_light} presents qualitative examples of the model's performance on the 3MDAD dataset under such variable lighting conditions. The model exhibits notable robustness in the middle and right panels, correctly classifying 'safe driving' despite severe sun glare and 'taking a picture' in bright ambient light. Conversely, the failure case in the left panel, which was misclassified as `Drinking', while the correct class was 'Smoking'. This illustrates how adverse lighting can increase the visual ambiguity of fine-grained actions.

Although our model shows resilience to lighting variations within its daytime training domain, this robustness does not extend to entirely different domains, such as nighttime driving. A common and effective engineering approach for creating an all-conditions, production-grade system involves a multi-modal strategy. Such a system would typically incorporate a light detection module to conditionally activate the appropriate model. A RGB-based model during daytime operation, and a complementary model optimized for the Near-Infrared (NIR) modality in low-light environments. Our work provides a highly optimized and crucial component for both scenarios within a comprehensive, all-conditions driver monitoring system.

\subsection{Discussion} 
Our contribution is a rich set of heatmap features for token selection, specifically designed for driver action recognition. These features consider factors such as object interaction and the static scale of the driver. This approach achieves a significant 30\% reduction in GFlops at the default settings while improving the accuracy over the baseline video transformers. Our comprehensive evaluation of various keep rate settings indicates that performance remains robust even when the computational cost is further reduced. Our model is well-suited for deployment on lower-end Jetson platforms and in real-world applications, 
setting a promising direction for research to make current video transformers usable for real-world applications.

\subsection{High risk classes analysis} 

\begin{figure}
    \centering
    \includegraphics[width=\linewidth]{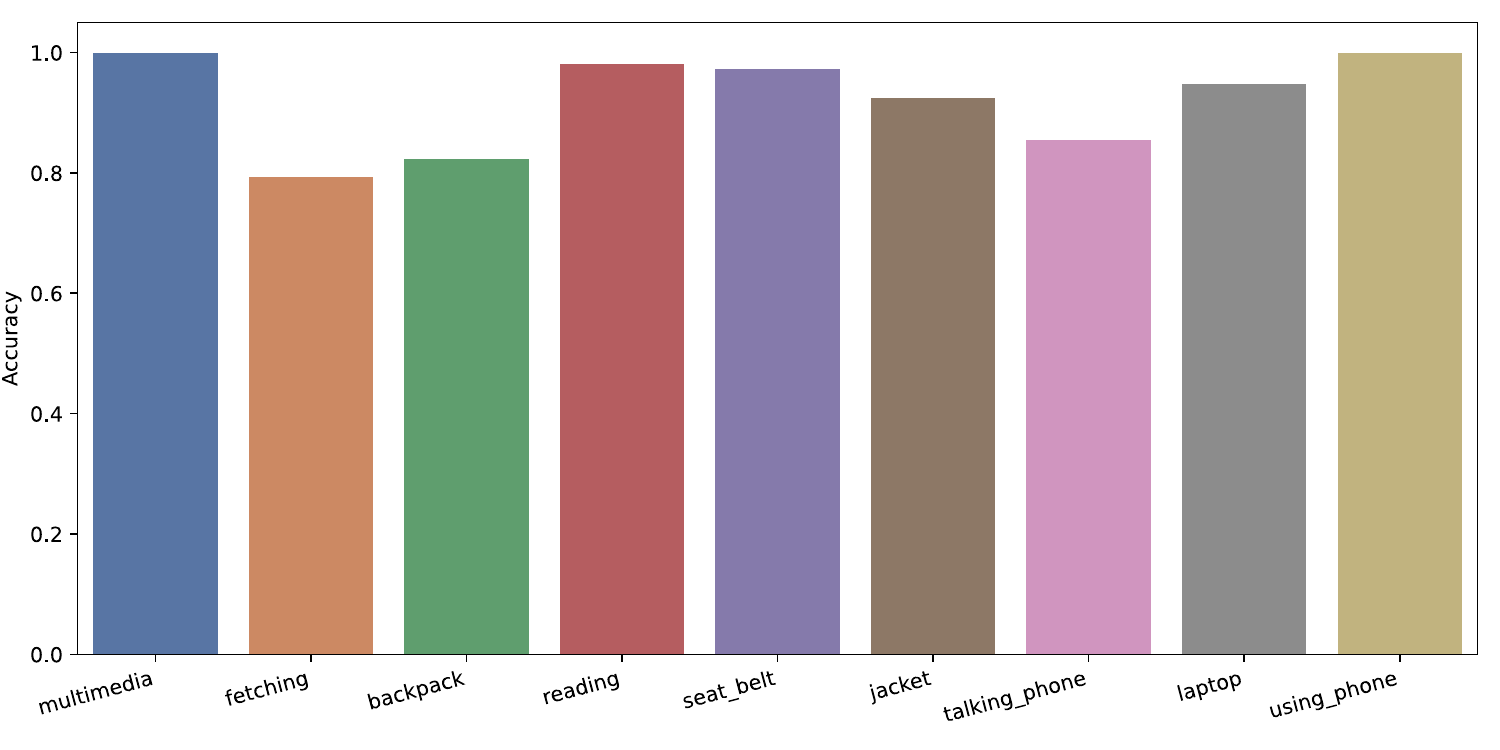}
    \caption{Accuracy breakdown for high-risk distraction categories on the Drive\&Act dataset. To provide a clearer analysis, fine-grained actions (e.g., 'opening laptop', 'closing laptop') were merged into broader semantic categories. The model demonstrates high robustness on critical distractions such as using a phone or interacting with the multimedia display, despite the token reduction.}
    \label{fig:high_risk_acc}
\end{figure}
To address the class imbalance inherent in the Drive\&Act dataset and evaluate \NuestroMetodo{} reliability in safety-critical scenarios, we analyze the performance on high-risk distraction categories in Fig.~\ref{fig:high_risk_acc}. For this analysis, we aggregated semantically similar fine-grained labels into broader categories. For example, 'opening', 'closing', and 'working on' a laptop were merged into 'laptop', and similar merging was applied to 'reading', 'jacket' and 'seat\_belt' interactions.

The results indicate that our method maintains a high accuracy on common distractions like 'using\_phone' and 'multimedia'. However, classes involving complex object manipulation, specifically 'fetching' and 'backpack', exhibit slightly lower accuracy. We observe that these two categories are frequently confused with one another because of the high similarity in their movements. Both actions typically involve the driver twisting their torso and extending an arm towards the passenger seat or rear cabin, creating visual ambiguity that is difficult to resolve without longer temporal context.

\subsection{Comparison with the state-of-the-art}

In this final section, we compare \NuestroMetodo{} with the state-of-the-art on three different driver action recognition datasets. Drive\&Act (Table~\ref{tab:driveactsota}), 100-Driver (Table~\ref{tab:hundredsota}) and 3MDAD (Table~\ref{tab:3mdadsota}). 

For the discussion of the Drive\&Act dataset (see Table~\ref{tab:driveactsota}), we refer to macro accuracy as it accounts for the class imbalance present in this dataset. Our model sets a new state-of-the-art. The effectiveness of our strategy is best demonstrated by the results against TransDARC~\cite{peng2022transdarc}, which uses the architecturally efficient Video Swin~\cite{liu2022videoswing} backbone. Our \NuestroMetodo{} achieves significantly higher accuracy (70.35\% vs. 55.30\%) with a lower computational cost (251 vs. 281 GFLOPs). This result suggests that for in this specialized domain, our semantically guided token selection is a more effective strategy than relying on a generically efficient backbone.

For a more direct comparison with \NuestroMetodo{}, we ensured parity in computational complexity by matching the number of GFlops in PO-GUISE. This comparison underscores the critical role of object information in this task. To evaluate \NuestroMetodo{} in another ViT-based backbone we have conducted additional experiments using InternVideo2-B/14~\cite{wang2024internvideo2} as the backbone. In these experiments, macro accuracy improves by approximately 1\% across all configurations. However, this comes at the cost of a 41\% increase in GFlops compared to the VideoMaeV2-base backbone. These results demonstrate that our \NuestroMetodo{} module can be successfully integrated into various ViT-based models, though its performance is inherently constrained by the architecture and layer composition of the chosen backbone.

\begin{table}
  \caption{Test results on Drive\&Act dataset in the front-top NIR camera. `*' denotes the methods reproduced using their original code and weight parameters. `\textbullet' denotes methods reproduced using MMAction2~\cite{mmaction2020}.
  }
  \setlength{\tabcolsep}{1mm}
  \footnotesize
  \centering
  \begin{tabular}{l|c|c|c}
    \toprule
    \multirow{2}{*}{Method} & Acc. & Macro Acc.. & GFlops\\
       & ($\uparrow$) & ($\uparrow$) &  ($\downarrow$) \\
    \hline
    st-MLP$^*$\cite{holzbock2022stmlp} & - & 33.51 & -\\
    Pose\cite{martin2019driveact}  & -  & 44.36 & -\\
    Interior\cite{martin2019driveact}  & -  & 40.30 & -\\
    2-stream\cite{martin2019driveact}  & -  & 45.39 & -\\
    3-stream\cite{martin2019driveact}  & -  & 46.95 & -\\
    I3D \textbullet\cite{carreira2017i3d}  & 71.50  & 48.87 & \first{33}\\
    SMOMS\cite{li2024smoms} & - & 51.39 & - \\
    OA-SAR\cite{li2023oasar} & - & 54.07 & - \\
    CTA-NET\cite{wharton2021ctanet}  & 65.25  & - & -\\
    3D-studentNet\cite{liu2023DADLightweight}  & 65.69  & - &  \second{37}\\
    Transdarc$^*$~\cite{peng2022transdarc}  & 66.92  & 55.30 & 281\\
    DRVMon-VM~\cite{pizarro2024drvmon}  & 77.27  & 62.64  & 360\\
    \hline
    VideoMAEv2-base & 83.50$\pm$ 2.0 & 68.27 $\pm$ 1.1
& 360\\
    PO-GUISE~\cite{pizarro2024poguise} & 83.60$\pm$ 3.1
 & 69.47$\pm$ 2.2& \third{251}\\ %
    \NuestroMetodo{} & \first{84.83$\pm$1.9} & \third{70.35$\pm$2.1} & \third{251}\\ %
    \hline
     InternVideo2-B/14 & \third{84.29$\pm$2.9} & 69.26$\pm$2.6& 509\\
    PO-GUISE~\cite{pizarro2024poguise} &\second{84.47$\pm$3.3} & \second{70.86$\pm$4.8} & 399\\ %
    \NuestroMetodo{} & 84.03$\pm$3.1 & \first{71.52$\pm$2.8} & 399\\ %
 
  \bottomrule
  \end{tabular}
  \label{tab:driveactsota}
\end{table}

In the context of the 100-Driver dataset, our model with VideoMAEv2 backbone sets new state-of-the-art results over previous approaches. While earlier methods relied on single-image models, our approach sets a new benchmark for video-based models on this dataset. Specifically, our proposed \NuestroMetodo{} surpasses the baseline VideoMAEv2-base by 2.24\% in accuracy.

\begin{table}
  \caption{Test results on 100-Driver dataset. `*' Indicates results obtained from \cite{wang2023100driver}.
  }
  \setlength{\tabcolsep}{1mm}
  \footnotesize
  \centering
  \begin{tabular}{l|c|c}
    \toprule
    \multirow{2}{*}{Method} & Acc.& GFlops\\
       & ($\uparrow$) &  ($\downarrow$) \\
    \hline
    MobileNetV3* & 76.0  & \first{0.22}\\
    SqueezeNet* & 79.6 & \third{1.72} \\
    EfficientNetB0* & 77.2 & \second{0.4} \\
    \hline
    VideoMAEv2-base & \third{91.30} & 360\\
    PO-GUISE~\cite{pizarro2024poguise} & \second{92.81} & 251\\ %
    \NuestroMetodo{} & \first{93.54} & 251\\ %
  \bottomrule
  \end{tabular}
  \label{tab:hundredsota}
\end{table}

For the 3MDAD dataset, our model with the VideoMAEv2 backbone also achieves state-of-the-art performance, as demonstrated in Table~\ref{tab:3mdadsota}. We surpass the previous best result set by MIFI~\cite{kuang2023MIFI}, a multi-view model. Specifically, our model improves accuracy by 9.52\% and reduces computational requirements by 28 GFLOPS compared to MIFI.

\begin{table}
  \caption{Test results on 3MDAD dataset (Acc. = averaged 5-fold test accuracy).
  }
  \label{tab:3mdadsota}
  \centering
  \begin{tabular}{l|c|c}
    \toprule
    Method & Acc. ($\uparrow$) & GFLOPS ($\downarrow$)\\
    \hline
    FPT\cite{wang2023FPT}  & 39.88 & \second{0.98} \\
    LW-transformer\cite{mohammed2024}  & 70.39& \first{0.329}  \\
    DADCNet\cite{su2023DADCNet} & 77.08 & \third{16} \\
    I3D-MIFI\cite{kuang2023MIFI} Single-view*  & 75.8 & 111 \\
    I3D-MIFI\cite{kuang2023MIFI} Multi-view*  & 83.9 & 223 \\
    \hline
    VideoMAEv2-base & \third{91.34$\pm$3.8} & 360 \\ 
    PO-GUISE\cite{pizarro2024poguise} & \second{92.71$\pm$2.4} & 251 \\
    \NuestroMetodo{} & \first{93.42$\pm$2.2} & 251 \\
  \bottomrule
  \end{tabular}
\end{table}

\section{Conclusions}

Current video transformers achieve state-of-the-art performance in action recognition but at the cost of quadratic complexity with respect to the number of input tokens. Techniques such as token selection have been employed to mitigate this computational demand, but they often discard critical information, particularly in specialized domains like driver action recognition.
Our proposed \NuestroMetodo{} effectively bridges the gap between general human action recognition models and the specialized requirements of driver action recognition. It achieves superior accuracy compared to previous methods, even in configurations with significantly reduced computational costs. In one of its most efficient settings, \NuestroMetodo{} reduces GFlops by 57\%, at an inference speed of 52 frames per second, while losing only 0.42\% accuracy compared to the baseline VideoMAEv2 model. Furthermore, our experiments demonstrate that the model is suitable for deployment onboard vehicles, particularly on Jetson hardware. Making such powerful models deployable on edge hardware is a critical step toward mitigating a primary cause of traffic incidents, thereby contributing to the broader goal of enhancing the safety and stability of the entire transportation system.

However, additional research is required to fully integrate \NuestroMetodo{} into a vehicle-ready system, including the incorporation of longer temporal contexts. This would necessitate transitioning to time-action localization tasks and utilizing datasets designed for such purposes. We plan to continue this line of research to have a fully functioning system to detect driver distraction. The models and code are available in \url{https://github.com/RicardoP0/poguise}.

\section*{Acknowledgements}

This work has been supported by projects PID2021-126623OB-I00, PID2022-137581OB-I00 and PID2024-161576OB-I00, funded by MICIU/AEI/10.13039/501100011033 and co-funded by the European Regional Development Fund (ERDF, “A way of making Europe”), by project PLEC2023-010343 (INARTRANS 4.0) funded by MCIN/AEI/10.13039/501100011033, and by the R\&D program TEC-2024/TEC-62 (iRoboCity2030-CM) and ELLIS Unit Madrid, granted by the Community of Madrid.

\bibliographystyle{IEEEtran}
\bibliography{actions}

\end{document}